\documentclass[conference]{IEEEtran}
\IEEEoverridecommandlockouts

\usepackage{cite}
\usepackage{amsmath,amssymb,amsfonts}
\usepackage{graphicx}
\usepackage{textcomp}
\usepackage{xcolor}
\usepackage{url}

\begin{document}

\title{Detection of Emotions in Hindi-English\\ Code-Mixed Text Data}

\author{\IEEEauthorblockN{Divyansh Singh}
\IEEEauthorblockA{\textit{Department of Computer Science and Engineering} \\
\textit{The LNMIIT}\\
Jaipur, India \\
18ucs127@lnmiit.ac.in}
\thanks{\textbf{Note on this revision.} An earlier version of this paper
stated that it was the first work to address emotion detection in
Hindi-English code-mixed text. That claim was incorrect. Vijay et al.
(NAACL-HLT Student Research Workshop, 2018) introduced an
emotion-annotated Hindi-English code-mixed corpus and a supervised
classification system for it, and Sasidhar et al. (Procedia Computer
Science, 2020) reported results on the same task. Section~II has been
rewritten to position this work with respect to theirs, and the
discussion, limitations and reference list have been revised accordingly.
No dataset or experimental result has changed.}}

\maketitle

\begin{abstract}
Hindi-English code-mixing, the alternation between the two languages
within a single utterance, accounts for a substantial share of
user-generated content in India. Because such text is written in the
Latin script without a standardised transliteration convention, one
underlying word surfaces in many spellings and many tokens fall outside
English lexica. This paper addresses emotion detection in
this setting as a four-way classification problem over anger, fear,
happiness and sadness, a finer-grained task than the polarity
classification targeted by most prior work. We make three
contributions. First, we construct a corpus of $1{,}589$ code-mixed
sentences drawn from Twitter and from video-platform comment sections,
annotated by two bilingual speakers with an inter-annotator agreement of
$\kappa = 0.94$; the video-comment portion covers a longer, less
hashtag-dominated register than the Twitter-only corpora used previously.
Second, we propose a normalisation procedure that clusters
transliteration variants by combining distributional similarity over
skip-gram vectors with a hard constraint on consonant identity, since
variation is carried almost entirely by vowels. Third, we compare five baselines under a common five-fold
cross-validation protocol: Na\"{i}ve Bayes over character and over word
$n$-grams, a word-level LSTM, a sub-word LSTM with convolved character
embeddings, and an SVM over frequency-based word vectors. The sub-word
LSTM attains the highest accuracy, $76.6\%$, against a majority-class
floor of $30.8\%$, and the controlled comparison against an otherwise
identical word-level LSTM isolates sub-word representation as the source
of that gain. On macro-averaged $F_1$, however, it ties with word
$n$-gram Na\"{i}ve Bayes at $0.77$, a margin that five-fold
cross-validation on $1{,}589$ instances cannot resolve. We report a
per-class error analysis and discuss the limitations imposed by corpus
scale.
\end{abstract}


\section{Introduction}

Code-mixing, the alternation between two or more languages within a
single conversation or utterance, is a routine communicative practice
in multilingual communities and is especially prevalent in computer-mediated
communication~\cite{sitaram2019}. In India, where a large fraction of
online interaction takes place on smartphones and social platforms,
Hindi-English code-mixing (often termed \textit{Hinglish}) constitutes a
substantial share of user-generated text. Because such text is written in
the Latin script without a standardised transliteration convention, it
presents difficulties for natural language processing pipelines designed
for either constituent language in isolation: tokens are absent from
English lexica, morphological analysers trained on Devanagari Hindi do
not apply, and the same underlying word appears under many surface forms.

Emotion detection is the natural language processing task of identifying
the affective state expressed in a span of text. It is distinct from,
and finer-grained than, polarity-oriented sentiment analysis: a
negative-polarity utterance may express anger, fear or sadness, and these
categories carry different implications for downstream applications such
as content moderation, conversational agents and public health
monitoring. Psychological accounts of emotion, notably Ekman's basic
emotion theory~\cite{ekman1992} and Plutchik's psychoevolutionary
model~\cite{plutchik1980}, provide the categorical label sets on which
most computational work is based.

While sentiment analysis in Hindi-English code-mixed text has received
considerable attention, culminating in shared tasks and benchmark
suites~\cite{patwa2020,khanuja2020}, the finer-grained problem of emotion
detection in this setting remains comparatively under-explored, and the
resources available for it are small. This paper contributes to that
smaller body of work. Specifically:

\begin{itemize}
\item We construct and release a corpus of $1{,}589$ Hindi-English
code-mixed sentences annotated with one of four emotion labels, drawn
from Twitter and from video-platform comments, with an inter-annotator
agreement of $\kappa = 0.94$.
\item We propose a normalisation procedure that clusters orthographic
variants arising from transliteration by combining distributional
similarity, computed over skip-gram vectors, with a hard constraint on
consonant identity.
\item We establish and compare five baselines spanning probabilistic,
kernel-based and neural approaches, and report a per-class error analysis
that identifies sentence length and class imbalance as the principal
sources of residual error.
\end{itemize}

The remainder of the paper is organised as follows. Section~\ref{sec:related}
situates the work within the literature on code-mixed NLP. Section~\ref{sec:dataset}
describes the corpus and its annotation. Section~\ref{sec:method} presents the
normalisation procedure, the task formulation and the baseline models.
Sections~\ref{sec:setup} and~\ref{sec:results} report the experimental setup and
results. Section~\ref{sec:limitations} discusses limitations, and
Section~\ref{sec:conclusion} concludes.

\section{Related Work}
\label{sec:related}

\subsection{Resources and Benchmarks for Code-Mixed NLP}

Computational work on code-switched language has been surveyed by Sitaram
et al.~\cite{sitaram2019}. Progress in the area has been driven largely by
the creation of shared tasks and annotated resources, beginning with the
first shared task on language identification in code-switched
data~\cite{solorio2014} and extending to part-of-speech tagging for
English-Hindi social media content~\cite{vyas2014}. For sentiment
specifically, SemEval-2020 Task~9 (SentiMix) released a Hinglish corpus of
approximately $20{,}000$ tweets annotated for word-level language
identity and sentence-level polarity, attracting $61$ participating teams
in the Hinglish track~\cite{patwa2020}. The GLUECoS
benchmark~\cite{khanuja2020} consolidates six code-switched tasks across
English-Hindi and English-Spanish and establishes that multilingual
models fine-tuned on code-switched data outperform cross-lingual
embedding approaches on most of them. The romanisation of Indic text has
additionally been addressed at the resource level: the Dakshina
dataset~\cite{roark2020} provides parallel Latin- and native-script text
together with romanisation lexicons for twelve South Asian languages,
offering a route to modelling transliteration variance that does not
depend on the code-mixed corpus itself.

\subsection{Sentiment Analysis in Hindi-English Code-Mixed Text}

Joshi et al.~\cite{joshi2016} introduced the sub-word LSTM for sentiment
analysis of Hindi-English code-mixed text, convolving character-level
embeddings to obtain sub-word representations that are robust to
orthographic variation. This architecture, illustrated in
Fig.~\ref{fig:subword}, is the direct antecedent of the strongest baseline
evaluated here. Subsequent systems submitted to SentiMix have shown that
BERT-style pre-trained encoders~\cite{devlin2019,conneau2020} and their
ensembles dominate the leaderboard for polarity
classification~\cite{patwa2020}.

\subsection{Emotion Detection in Hindi-English Code-Mixed Text}

The finer-grained emotion detection task has a shorter history. Vijay et
al.~\cite{vijay2018} presented the first Hindi-English code-mixed corpus
annotated for emotion, comprising tweets labelled with the expressed
emotion together with word-level source-language and causal-language
annotations, and evaluated a supervised classification system over it.
Sasidhar et al.~\cite{sasidhar2020} subsequently applied a CNN-BiLSTM
architecture to the same problem, reporting that a convolutional layer
placed ahead of the recurrent layer both improves accuracy and reduces
training cost. The task is thus supported by only a small number of
corpora, none of them large.

The present work is therefore not the first to address emotion detection
in this setting. It differs from the prior literature in three respects.
First, our corpus is drawn in part from video-platform comment sections
rather than exclusively from Twitter, and therefore covers a register with
longer and less hashtag-dominated utterances. Second, we address
transliteration variance explicitly through a constrained clustering
procedure applied prior to modelling, rather than relying solely on the
model to absorb it. Third, we provide a controlled comparison between
word-level and sub-word-level representations under an identical
cross-validation protocol, which isolates the contribution of sub-word
modelling on a corpus of this size.

\section{Dataset}
\label{sec:dataset}

Following the recommendations of Bender and Friedman~\cite{bender2018},
we describe the provenance, the annotator population and the annotation
protocol of the corpus in this section.

\subsection{Data Collection}

Candidate sentences were collected using data-scraping tools applied to
Twitter and to the comment sections of video-streaming platforms. Each
data point consists of a single sentence written in Hindi-English
code-mixed language, transliterated into the Latin script, together with
one emotion label. Retweets, duplicates and sentences containing no Hindi
token were discarded. All records were stripped of user handles, URLs and
other directly identifying metadata prior to annotation; no user-level
information was retained.

\subsection{Label Set and Annotation Protocol}

We adopt a four-way label set, namely Angry (A), Fear (F), Sad (S) and
Happy (H), corresponding to four of Ekman's six basic
emotions~\cite{ekman1992}.
Disgust and surprise were excluded because they were too sparsely
attested in the collected data to support reliable estimation, a
restriction consistent with the label sets used in related work on
code-mixed emotion~\cite{vijay2018,sasidhar2020}.

The corpus was annotated independently by two annotators, both native
speakers of Hindi and fluent in English. Annotators labelled each
sentence with the emotion they judged the author to be expressing.
Inter-annotator agreement, measured as Cohen's
$\kappa$~\cite{mchugh2012,artstein2008}, was $0.94$, which falls in the
range conventionally described as almost perfect agreement.

Of the $5{,}986$ sentences initially collected, $1{,}589$ ($26.5\%$)
were retained in the final corpus; the remainder were discarded during
filtering and annotation. The composition of the resulting corpus is reported in
Table~\ref{tab:composition}, and representative examples are given in
Table~\ref{tab:examples}. The distribution is moderately imbalanced: the
majority class (Happy) accounts for $30.8\%$ of instances and the
minority class (Fear) for $19.1\%$, so a majority-class classifier would
achieve $30.8\%$ accuracy. All results in Section~\ref{sec:results}
should be read against this floor.

\begin{table}[htbp]
\caption{Composition of the Dataset}
\begin{center}
\begin{tabular}{|l|c|c|}
\hline
\textbf{Class (Emotion)} & \textbf{Instances} & \textbf{Share (\%)} \\
\hline
Angry & 471 & 29.6 \\
\hline
Fear  & 304 & 19.1 \\
\hline
Sad   & 324 & 20.4 \\
\hline
Happy & 490 & 30.8 \\
\hline
\textbf{Total} & \textbf{1589} & \textbf{100.0} \\
\hline
\end{tabular}
\label{tab:composition}
\end{center}
\end{table}

\begin{table}[htbp]
\caption{Examples of Annotated Data Points}
\begin{center}
\begin{tabular}{|l|c|}
\hline
\textbf{Sentence} & \textbf{Label} \\
\hline
Kutte chup reh tu & A \\
\hline
Aaaj mei bahut khushh hu & H \\
\hline
Bhoot bhoot bachao mujhe & F \\
\hline
Mujhe bohot dukh hai RIP & S \\
\hline
\end{tabular}
\label{tab:examples}
\end{center}
\end{table}

\subsection{Availability}

The corpus is available from the author on request. Consistent with
platform terms of service, no user-level metadata is redistributed.

\section{Methodology}
\label{sec:method}

\subsection{Normalising Transliteration Variants}

Transliteration from Hindi into the Latin script is not orthographically
standardised, so a single Hindi word maps to many English surface forms.
The word \textit{hain}, for instance, is variously written \textit{hai},
\textit{ha} or \textit{h}; the word \textit{khoobsurat} appears as
\textit{khbsrt}, \textit{khubsurat} or \textit{khoobsoorat}. Two
regularities are apparent across such variant sets. First, the consonant
skeleton is preserved, since variation is carried almost entirely by
vowels and by vowel elision. Second, because the variants denote the same
word and discharge the same function, they occur in near-identical
contexts and therefore occupy neighbouring regions of a distributional
semantic space.

We exploit both regularities jointly. Skip-gram vectors~\cite{mikolov2013}
are trained over the unlabelled portion of the scraped data, and words
are clustered under the similarity measure
\begin{equation}
f(v_1, v_2)=
\begin{cases}
\cos(v_1, v_2), & \text{if } c(v_1) = c(v_2),\\
0, & \text{otherwise},
\end{cases}
\label{eq:sim}
\end{equation}
where $c(\cdot)$ returns the ordered consonant sequence of a word and
$\cos(\cdot,\cdot)$ denotes cosine similarity between the corresponding
skip-gram vectors. The consonant constraint acts as a hard filter that
prevents distributionally similar but etymologically unrelated words from
being merged; the cosine term then ranks the surviving candidates. Words
are agglomerated into a 
cluster when $f(v_1,v_2)$ exceeds a fixed similarity threshold, and
every member of a
cluster is replaced by the most frequent member. This substitution rests
on the assumption that the highest-frequency form is the most likely to
be the intended one, which holds when the corpus is large enough for the
frequency ranking to be stable; we return to this assumption in
Section~\ref{sec:limitations}. The procedure reduces the number of
distinct types in the vocabulary by collapsing each variant set onto a
single form.

An alternative to explicit normalisation is to adopt a representation
that is inherently robust to sub-word variation, such as fastText
embeddings with character $n$-gram composition~\cite{bojanowski2017} or
the character-level convolution of Joshi et al.~\cite{joshi2016}; we
compare against the latter directly in Section~\ref{sec:results}.

\subsection{Task Formulation}

The task is a single-label multiclass classification problem in which the
input is a code-mixed sentence and the output is a label
$y \in \{F, A, S, H\}$. Writing $y_c$ for the one-hot encoding of the
gold label and $p_c(x)$ for the model's predicted probability of class
$c$ given input $x$, the models are trained to minimise the categorical
cross-entropy~\cite{zhang2018}
\begin{equation}
\mathcal{L}(x, y) = -\sum_{c \in \{F,A,S,H\}} y_c \log p_c(x).
\label{eq:loss}
\end{equation}

\begin{figure}[htbp]
\centerline{\includegraphics[width=\columnwidth]{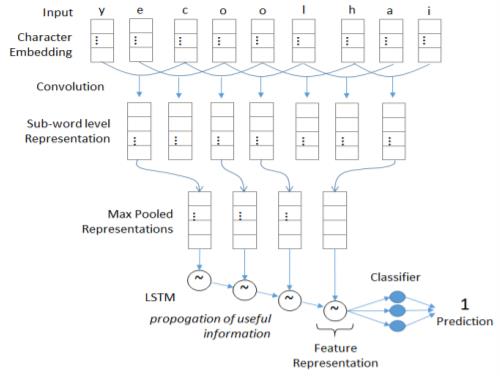}}
\caption{Illustration of the Sub-word LSTM architecture, reproduced
from Joshi et al.~\cite{joshi2016}.}
\label{fig:subword}
\end{figure}

\subsection{Baseline Models}

Five baselines were trained and evaluated under the formulation above.
They were chosen to span three modelling families (probabilistic,
kernel-based and neural) and, within the neural family, to contrast
word-level with sub-word-level representations.

\begin{itemize}
\item \textbf{Na\"{i}ve Bayes over character $n$-grams.} A multinomial
Na\"{i}ve Bayes classifier~\cite{zhang2007} over character $n$-gram
counts. Among the orders searched, $n=8$ gave the best cross-validated
accuracy.

\item \textbf{Na\"{i}ve Bayes over word $n$-grams.} The same classifier
over word $n$-grams. A combined unigram-and-bigram feature space
outperformed either order used alone.

\item \textbf{Word-level LSTM.} A single-layer Long Short-Term Memory
network~\cite{hochreiter1997} over word2vec
embeddings~\cite{mikolov2013} learned jointly with the recurrent layer.

\item \textbf{Sub-word LSTM.} The architecture of Joshi et
al.~\cite{joshi2016}, in which character-level embeddings are passed
through a one-dimensional convolution and max-pooled into sub-word
representations before being consumed by the LSTM
(Fig.~\ref{fig:subword}). This is a code-mixed adaptation of the sentence
classification architecture of Kim~\cite{kim2014}.

\item \textbf{Support vector machine.} An SVM~\cite{hearst1998} over
vectorised sequences constructed from frequency-distribution word
vectors.
\end{itemize}

\section{Experimental Setup}
\label{sec:setup}

All models were trained end to end and evaluated under five-fold
cross-validation, with folds stratified by class so that the label
distribution of Table~\ref{tab:composition} is preserved in every fold.
We report per-class $F_1$, macro-averaged $F_1$ and accuracy. Macro-averaging
is reported because accuracy alone is misleading on an imbalanced corpus:
it rewards performance on the two larger classes disproportionately.

The neural models were trained with the Adam optimiser and early
stopping. Reported figures are the mean over the five folds.

\section{Results and Discussion}
\label{sec:results}

Table~\ref{tab:results} reports the cross-validated results. The sub-word
LSTM attains the highest accuracy, $76.6\%$, exceeding the majority-class
floor of $30.8\%$ by a wide margin and the next-best model, word $n$-gram
Na\"{i}ve Bayes, by $1.5$ accuracy points.

\begin{table}[htbp]
\caption{Class-wise $F_1$, Macro-$F_1$ and Accuracy under Five-Fold Cross-Validation}
\begin{center}
\footnotesize
\setlength{\tabcolsep}{3pt}
\begin{tabular}{|l|c|c|c|c|c|c|}
\hline
\textbf{Baseline} & \textbf{Sad} & \textbf{Angry} & \textbf{Happy} &
\textbf{Fear} & \textbf{Macro} & \textbf{Acc.} \\
& & & & & \textbf{$F_1$} & \textbf{(\%)} \\
\hline
Majority class & 0.00 & 0.00 & 0.47 & 0.00 & 0.12 & 30.8 \\
\hline
Na\"{i}ve Bayes, char $n$-grams & 0.70 & 0.73 & 0.68 & \textbf{0.82} & 0.73 & 72.3 \\
\hline
Na\"{i}ve Bayes, word $n$-grams & 0.73 & 0.76 & 0.77 & 0.81 & 0.77 & 75.1 \\
\hline
LSTM (word2vec) & 0.72 & 0.68 & 0.72 & \textbf{0.82} & 0.74 & 72.6 \\
\hline
Sub-word LSTM & \textbf{0.79} & \textbf{0.78} & \textbf{0.78} & 0.73 & \textbf{0.77} & \textbf{76.6} \\
\hline
SVM & 0.68 & 0.66 & 0.70 & \textbf{0.82} & 0.72 & 70.1 \\
\hline
\end{tabular}
\label{tab:results}
\end{center}
\end{table}

\subsection{Sub-Word versus Word-Level Representations}

The clearest finding is the gap between the word-level LSTM and the
sub-word LSTM, which share a recurrent backbone and differ only in the
representation supplied to it. The sub-word variant improves accuracy by
$4.0$ points and macro-$F_1$ by $0.03$. We attribute this to residual
transliteration variance: the clustering procedure of
Section~\ref{sec:method} merges variants that are attested often enough
for their skip-gram vectors to be reliably estimated, but rare variants
remain unmerged and are mapped to out-of-vocabulary tokens by the
word-level model. Character-level embeddings sidestep this failure mode
because they never commit to a word-level vocabulary in the first place.
This is consistent with the original motivation for the
architecture~\cite{joshi2016} and with the gains reported for
character-composed representations more
generally~\cite{bojanowski2017,roark2020}.

\subsection{Strength of the Probabilistic Baselines}

Na\"{i}ve Bayes over word $n$-grams is competitive with the sub-word
LSTM: it trails by $1.5$ accuracy points, and the two models are tied on
macro-$F_1$ at $0.77$. Given a corpus of $1{,}589$ instances and a
five-fold protocol, each test fold contains roughly $318$ sentences, and
a difference of this magnitude falls within the range that
cross-validation on a corpus of this size cannot resolve. We therefore do
not claim a significant advantage for the sub-word LSTM on macro-$F_1$;
establishing one would require a paired significance test over the folds,
with the multiple-comparison correction appropriate to five
systems~\cite{demsar2006,dror2018}.

That a bag-of-$n$-grams model performs this well is itself informative.
Emotion in short social media utterances is frequently carried by one or
two lexical items, such as an intensifier, an interjection or an
expletive, and a model that conditions independently on such items captures most of
the available signal. The margin available to sequence models is
correspondingly narrow at this corpus size.

\subsection{Per-Class Behaviour}

The Fear class is the smallest in the corpus yet is the best-classified
class for three of the five models, reaching $F_1 = 0.82$ for the
character $n$-gram, word-level LSTM and SVM baselines. This suggests that
fear is expressed through a comparatively closed and distinctive lexicon
(\textit{bachao}, \textit{darr}, \textit{bhoot}), which surface-feature
models capture readily. The sub-word LSTM is the exception, scoring
$0.73$ on Fear while leading on the other three classes: it gains on the
classes that require broader contextual discrimination and loses on the
one that is best served by memorising a small lexicon, which is the
behaviour expected of a higher-capacity model on the smallest class.

\section{Limitations}
\label{sec:limitations}

\textbf{Corpus size.} At $1{,}589$ instances the corpus is small by the
standards of emotion detection generally and of code-mixed sentiment
resources specifically; SentiMix, by comparison, released roughly
$20{,}000$ Hinglish tweets~\cite{patwa2020}. Small test folds inflate the
variance of the reported figures and limit the resolution of any
comparison between models.

\textbf{Sentence length.} Sub-word and character-level $n$-gram models
degrade on long sentences, where the emotional cue is diluted across a
longer character sequence and the max-pooling operation discards
positional information.

\textbf{Assumptions in the normalisation procedure.} The substitution
rule of Section~\ref{sec:method} assumes that the most frequent member of
a cluster is the correct form. On a corpus of this size the frequency
ranking within a cluster is itself noisy, and the consonant-identity
constraint will fail for variants that differ in consonant gemination or
in the aspirated/unaspirated distinction (\textit{khushi} versus
\textit{kushi}), which are precisely the distinctions that Latin-script
transliteration renders inconsistently.

\textbf{Absence of pre-trained encoders.} The baselines omit the
transformer encoders that have come to define the state of the art on
code-mixed tasks~\cite{devlin2019,conneau2020}, and which dominated the
SentiMix leaderboard~\cite{patwa2020}. The comparison presented here is
therefore internally controlled but does not situate these models against
contemporary systems.

\textbf{Domain and population.} The corpus is drawn from two platforms
and annotated by two annotators from a single linguistic and regional
background. Emotion annotation is known to be sensitive to annotator
background, and the high $\kappa$ reported here reflects agreement
between two similar annotators rather than the reproducibility of the
labels across the broader population of Hinglish speakers.

\section{Ethical Considerations}

The data were collected from publicly accessible posts and comments.
User handles, URLs and identifying metadata were removed before
annotation, and no attempt was made to link records to individuals or to
infer user-level attributes. Emotion classifiers of this kind can be
applied to the affective profiling of individuals without their
knowledge; we release the corpus for research on code-mixed language
understanding and note that its small size and narrow platform coverage
make it unsuitable for deployment in any consequential decision-making
setting.

\section{Conclusion and Future Work}
\label{sec:conclusion}

We have presented a corpus of $1{,}589$ Hindi-English code-mixed
sentences annotated for four emotions, a constrained clustering procedure
for normalising transliteration variants, and a controlled comparison of
five baseline classifiers. The sub-word LSTM achieves the highest
accuracy, $76.6\%$, and the comparison against an otherwise identical
word-level LSTM isolates sub-word representation as the source of that
gain. A word $n$-gram Na\"{i}ve Bayes model remains competitive, which we
read as a consequence of emotion being lexically concentrated in short
utterances.

The most direct route to improvement is a larger corpus: every limitation
identified in Section~\ref{sec:limitations} is either caused or
aggravated by scale. Beyond this, three extensions are natural.
Fine-tuning multilingual pre-trained encoders on code-switched
data~\cite{devlin2019,conneau2020}, which GLUECoS~\cite{khanuja2020}
identifies as the strongest general-purpose approach in this setting,
would establish how much of the remaining error is attributable to
representation rather than to data; romanisation resources such as
Dakshina~\cite{roark2020} offer a complementary route to the same end.
Attention-based architectures~\cite{vaswani2017} would address the
degradation observed on long sentences, at a computational cost that must
be weighed against the modest scale of the task. Finally, evaluation on
the emotion-annotated corpus of Vijay et al.~\cite{vijay2018} would allow
the normalisation procedure proposed here to be assessed independently of
the corpus on which it was developed.

\end{document}